\documentclass[10pt,twocolumn,letterpaper]{article}

\usepackage{cvpr}
\usepackage{epsfig}
\usepackage{graphicx}
\usepackage{amsmath}
\usepackage{amssymb}
\usepackage{multirow}
\usepackage{authblk,url}
\renewcommand\footnotemark{}

\usepackage{caption}
\usepackage{sidecap}

\cvprfinalcopy %

\ifcvprfinal\fi
\begin{document}

\title{An end-to-end TextSpotter with Explicit Alignment and Attention\thanks{Appearing in Proc.\ IEEE
    Conf. Computer Vision and Pattern Recognition, 2018.}
}

    \author{Tong He$^{1,*}$, Zhi Tian$^{1,*}$, Weilin Huang$^{3}$, Chunhua
        Shen$^{1,\dagger}$\thanks{The first two authors contribute equally. C. Shen is the
        corresponding author (e-mail: chunhua.shen@adelaide.edu.au).}, Yu Qiao$^4$, Changming Sun$^{2}$\\
    $^1$University of Adelaide, Australia    $^{2}$Data61, CSIRO, Australia   $^3$Malong Technologies\\
    $^4${Shenzhen Institutes of Advanced Technology, Chinese Academy of Sciences} \\
}

\thispagestyle{empty}

\maketitle

\thispagestyle{empty}

\begin{abstract}
\vspace{-0.3cm}
Text detection and recognition in natural images have long been considered as two separate tasks that are processed sequentially.
  Training of two tasks in a unified framework is non-trivial due to significant differences in
  %learning difficulties and convergence rates.
  optimisation difficulties.
In this work, we present a conceptually simple yet efficient framework that simultaneously processes the two tasks in one shot.
Our main contributions are three-fold:
  1) we propose a novel \emph{text-alignment} layer that allows it to precisely compute convolutional features of a text instance in arbitrary orientation, which is the key to boost the performance;
  2) a character attention mechanism is introduced by using character spatial information as explicit supervision, leading to large improvements in recognition;
  3) two technologies, together with a new RNN branch for word recognition, are integrated seamlessly into a single model which is end-to-end trainable. This allows the two tasks to work collaboratively by sharing convolutional features, which is critical to identify challenging text instances.
Our model achieves impressive results in end-to-end recognition on the ICDAR2015 \cite{ICDAR2015} dataset,
  significantly advancing  most recent results \cite{Busta2017}, with improvements of F-measure from $(0.54, 0.51, 0.47)$
  to $(0.82, 0.77, 0.63)$,
  by using a strong, weak and generic lexicon respectively.
  Thanks to joint training, our method can also serve as a good detector by achieving a new state-of-the-art detection performance on
  two  datasets.
\vspace{-0.5cm}
\end{abstract}

\tableofcontents
\clearpage

\renewcommand{\vspace}[1]{{}}

	\section{Introduction}
\vspace{-0.2cm}
The goal of text spotting is to map an input natural image into a set of character sequences or word transcripts and corresponding location. It has attracted increasing attention in the vision community, due to its numerous potential applications. It has made rapid progress riding on the wave of recent deep learning technologies, as substantiated by recent works \cite{Jaderberg2015, HeP2017, Busta2017, Li2017, HeW2017, Tian2016, Zhou2017, Shi2017, Tian2017, Liu2018}. However, text spotting in the wild still remains an open problem, since text instances often exhibit vast diversity in font, scale and orientation with various illumination affects, which often come with a highly complicated background.

\begin{figure*}
	\begin{center}
        \includegraphics[width=.6\textwidth]{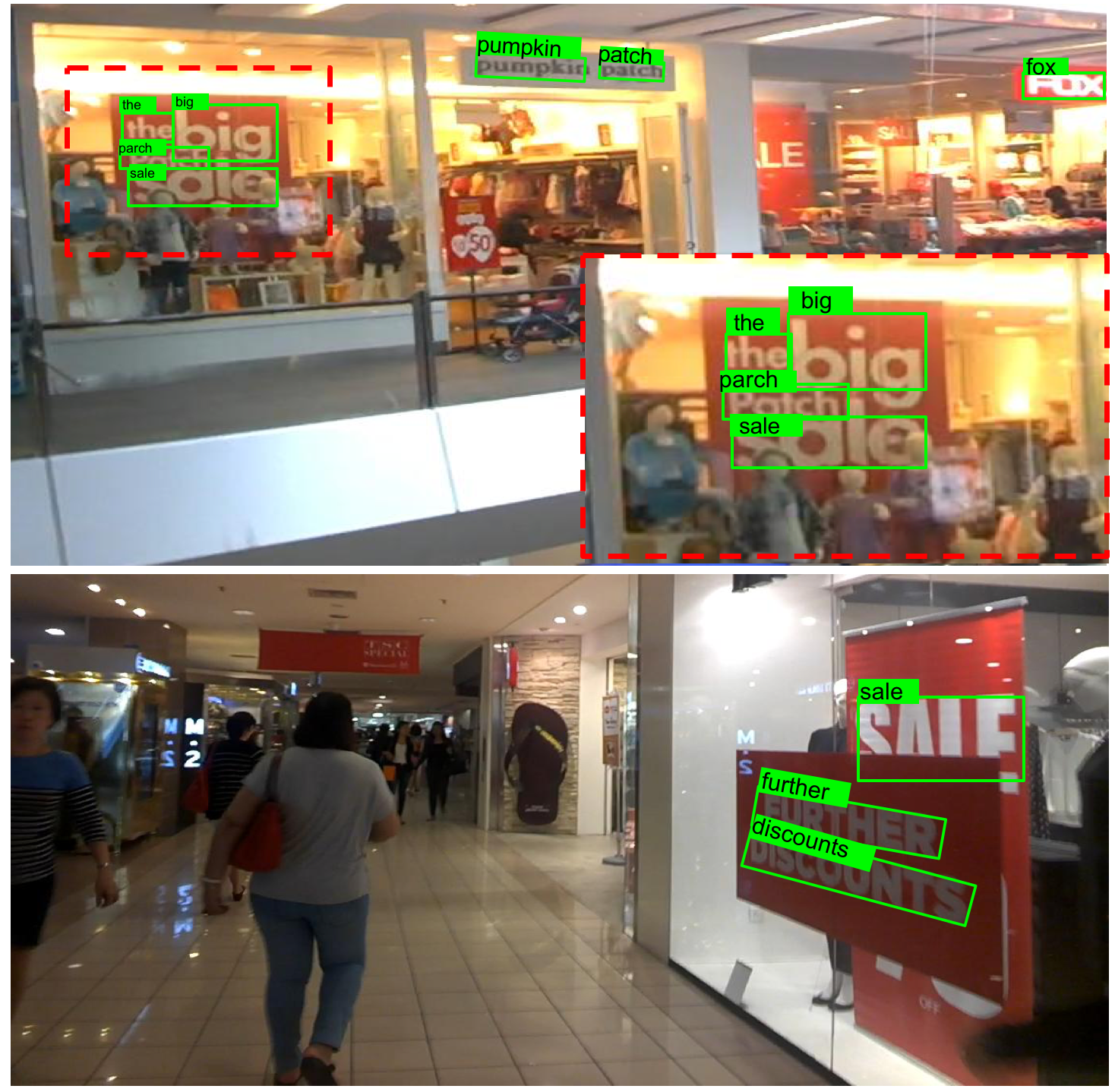}
		\vspace{-0.2cm}
	\end{center}
	\caption{Illustrations of the results on ICDAR 2015 by our proposed method, which can detect all possible text regions and recognize relevant transcriptions in just one shot.}
	\label{fig:main}
	\vspace{-0.65cm}
\end{figure*}
Past works in text spotting often consider it as two individual tasks: text detection and word recognition, which are implemented sequentially. The goal of text detection is to precisely localize all text instances (e.g., words) in a natural image, and then a recognition model is processed repeatedly through all detected regions for recognizing corresponding text transcripts. Recent approaches for text detection are mainly extended from general object detectors (such as Faster R-CNN \cite{Ren2015} and SSD \cite{LiuW2016}) by directly regressing a bounding box for each text instance, or from semantic segmentation methods (e.g., Fully Convolutional Networks (FCN) \cite{Long2015}) by predicting a text/non-text probability at each pixel. With careful model design and development, these approaches can be customized properly towards this highly domain-specific task, and achieve the state-of-the-art performance \cite{HeP2017, HeW2017, Tian2016, Zhou2017, Shi2017, Zhang2016}. The word recognition can be cast into a sequence labeling problem where convolutional recurrent models have been developed recently \cite{Shi2017, HeP2016}. Some of them were further incorporated with an attention mechanism for improving the performance \cite{Lee2016, Bahdanau2016}. However, training two tasks separately does not exploit the full potential of convolutional networks, where the convolutional features are not shared. It is natural for us to make a more reliable decision if we clearly understand or recognize the meaning of a word and all characters within it. Besides, it is also possible to introduce a number of heuristic rules and hyper-parameters that are costly to tune, making the whole system highly complicated.

Recent Mask R-CNN \cite{HeK2017} incorporates an instance segmentation task into the Faster R-CNN \cite{Ren2015} detection framework, resulting in a multi-task learning model that jointly predicts a bounding box and a segmentation mask for each object instance. Our work draws inspiration from this pipeline, but has a different goal of learning a direct mapping between an input image and a set of character sequences. We create a recurrent sequence modeling branch for word recognition within a text detection framework, where the RNN based word recognition is processed in parallel to the detection task.

However, the RNN branch, where the gradients are back-propagated through time, is clearly much more difficult to optimize than the task of bounding box regression in detection. This naturally leads to significant differences in learning difficulties and convergence rates between two tasks, making the model particularly hard to be trained jointly.
For example, the magnitude of images for training a text detection model is about $10^3$ (e.g., 1000 training images in the ICDAR 2015 \cite{ICDAR2015}) , but the number is increased significantly by many orders of magnitude when a RNN based text recognition model is trained, such as the 800K synthetic images used in \cite{Gupta2016}.  Furthermore, simply using a set of character sequences as direct supervision may be too abstractive (high-level) to provide meaningful detailed information for training such an integrated model effectively, which will make the model difficult to convergence. In this work, we introduce strong spatial constraints in both word and character levels, which allows the model to be optimized gradually by reducing the search space at each step.

\textbf{Contributions}
In this work, we present a single-shot textspotter capable of learning a direct mapping between an input image and a set of character sequences or word transcripts. We propose a solution that combines a \emph{text-alignment} layer tailed for multi-orientation text detection, together with a character attention mechanism that explicitly encodes strong spatial information of characters into the RNN branch, as shown in Fig. \ref{fig:main}. These two technologies faithfully preserve the exact spatial information in both text instance and character levels, playing a key role in boosting the overall performance. We develop a principled learning strategy that allows the two tasks to be trained collaboratively by sharing convolutional features. Our main contributions are described as follows.

Firstly, we develop a text-alignment layer by introducing a grid sampling scheme instead of conventional RoI pooling. It computes fixed-length convolutional features that precisely align to a detected text region of arbitrary orientation, successfully avoiding the negative effects caused by orientation changing and quantization factor of the RoI pooling.

Secondly, we introduce a character attention mechanism by using character spatial information as an addition supervision. This explicitly encodes strong spatial attentions of characters into the model, which allows the RNN to focus on current attentional features in decoding, leading to performance boost in word recognition.

Thirdly, both approaches, together with a new RNN branch for word recognition, are integrated elegantly into a CNN detection framework, resulting in a single model that can be trained in an end-to-end manner. We develop a principled and intuitive learning strategy that allows the two tasks to be trained effectively by sharing features, with fast convergence.

Finally, we show by experiments that word recognition can significantly improve detection accuracy
in our model, demonstrating strong complementary nature of them, which is unique to this highly
domain-specific application. Our model achieves new state-of-the-art results on the ICDAR2015 in
end-to-end recognition of multi-orientation texts, largely outperforming the most recent results in
\cite{Busta2017}, with improvements of F-measure from $(0.54, 0.51, 0.47)$ to $(0.82, 0.77, 0.63)$ in terms of using a strong, weak and generic lexicon.
Code is available at
\url{https: //github.com/tonghe90/textspotter}

\begin{figure*}
	\begin{center}
		\includegraphics[width=12.5cm]{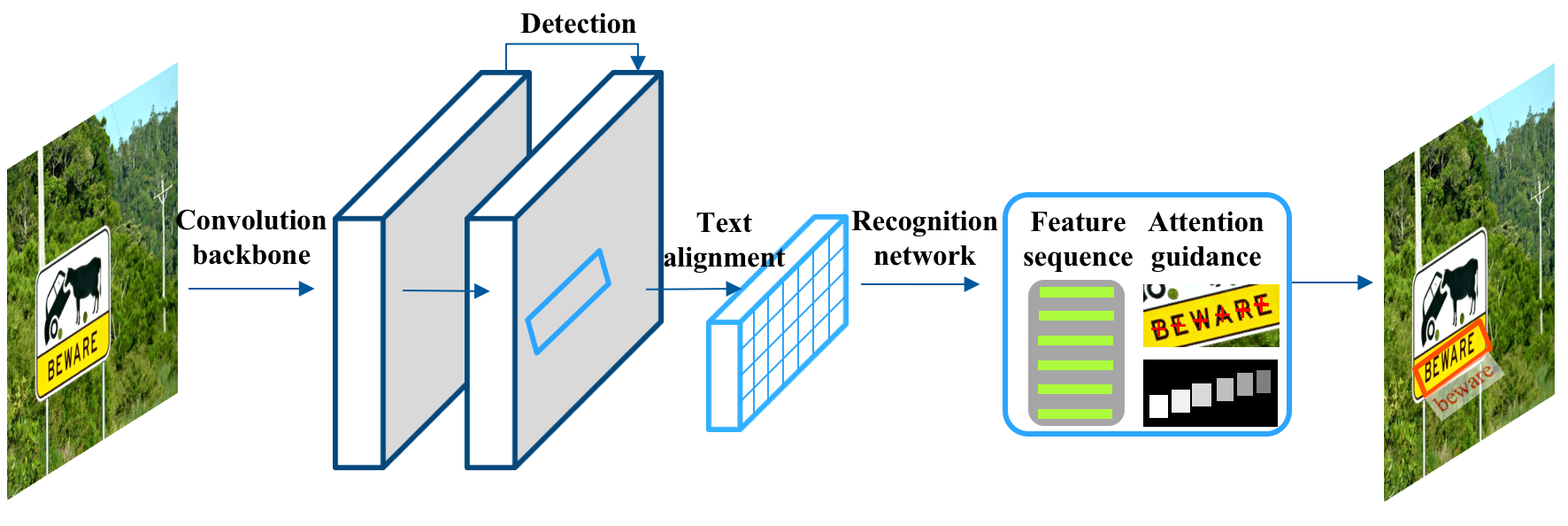}
	\end{center}
	\vspace{-0.5cm}
	\caption{The framework of our method. \emph{text-alignment} layer is proposed to extract
    accurate sequence features within a detected quadrilateral of multi-orientation. A novel
character attention mechanism is applied
to guide the decoding process with explicit supervision. The whole framework can be trained in an end-to-end manner.}
	\label{fig:net}
	\vspace{-0.5cm}
\end{figure*}

	\textbf{Related work}
Here we  briefly introduce some related works on text detection,  recognition and end-to-end wordspotting.

\textit{Scene text detection} 
Recently, some methods cast previous character based detection \cite{Huang2013, Huang2014, HeT2016, Yao2012} into direct text region estimation \cite{Liao2017, Zhou2017, Zhang2016, HeT2016b, HeP2017, Zhang2015, Yao2016}, avoiding multiple bottom-up post-processing steps by taking word or text-line as a whole. Tian~\etal \cite{Tian2016} modified Faster-RCNN \cite{Ren2015} by applying a recurrent structure on the convolution feature maps of the top layer horizontally. The methods in \cite{HeP2017, Liao2017} were inspired from \cite{LiuW2016}. They both explored the framework from generic objects and convert to scene text detection by adjusting the feature extraction process to this domain-specific task. However, these methods are based on prior boxes, which need to be carefully designed in order to fulfill the requirements for training.
Methods of direct regression for inclined bounding boxes, instead of offsets to fixed prior boxes, have been proposed recently. EAST \cite{Zhou2017} designed a fully convolutional network structure which outputs a pixel-wise prediction map for text/non-text and five values for every point of text region, i.e., distances from the current point to the four edges with an inclined angle. He~\etal \cite{HeW2017} proposed a method to generate arbitrary quadrilaterals by calculating offsets between every point of text region and vertex coordinates.

\textit{Scene text recognition} With the success of recurrent neural networks on digit recognition and speech translation, a lot of works have been proposed for text recognition. He~\etal \cite{HeP2016} and Shi~\etal \cite{Shi2017,Shi2016} treat text recognition as a sequence labeling problem by introducing LSTM \cite{Alex2014} and connectionist temporal classification (CTC) \cite{Alex2009} into a unified framework. \cite{Lee2016} proposed an attention-based LSTM for text recognition, which mainly contains two parts: encoder and decoder. In the encoding stage, text images are transformed into a sequence of feature vectors by CNN/LSTM. Attention weights, indicating relative importance for recognition, will be learned during the decoding stage. However, these weights are totally learned by the distribution of data and no supervision is provided to guide the learning process.

\textit{End-to-end wordspotting}
End-to-end wordspotting is an emerging research area. Previous methods usually try to solve it by splitting the whole process into two independent problems: training two cascade models, one for detection and one for recognition.
Detected text regions are firstly cropped from original image, followed by affine transforming and rescaling. Corrected images are repeatedly precessed by recognition model to get corresponding transcripts.
However, training errors will be accumulated due to cascading models without sharable features. Li~\etal \cite{Li2017} proposed a unified network that simultaneously localizes and recognizes text in one forward pass by sharing convolution features under a curriculum strategy. But the existing RoI pooling operation limits it to detect and recognize only horizontal examples.
Busta~\etal \cite{Busta2017} brought up deep text spotter, which can solve wordspotting of multi-orientation problem. However, the method does not have sharable feature, meaning that the recognition loss of the later stage has no influence on the former localization results.

	\section{Single Shot TextSpotter by Joint Detection and Recognition}
\vspace{-0.2cm}
In this section, we present the details of the proposed textspotter which learns a direct mapping between an input image and a set of word transcripts with corresponding bounding boxes of arbitrary orientations. Our model is a fully convolutional architecture built on the PVAnet framework \cite{Hong2016}. As shown in Fig. \ref{fig:net}, we introduce a new recurrent branch for word recognition, which is integrated into our CNN model in parallel with the existing detection branch for text bounding box regression.
The RNN branch is composed of a new \emph{text-alignment} layer and a LSTM-based recurrent module with a novel character attention embedding mechanism.
The \emph{text-alignment} layer extracts precise sequence feature within the detected region, preventing encoding irrelevant texts or background information. The character attention embedding mechanism regulates the decoding process by providing more detailed supervisions of characters.
Our textspotter directly outputs final results in one shot, without any post-processing step except for a simple non-maximum suppression (NMS).

\begin{figure*}
	\begin{center}
		\includegraphics[width=11.5cm]{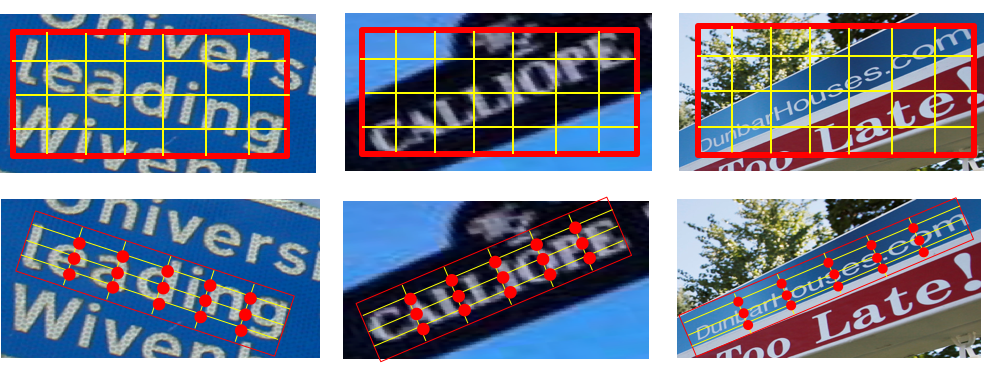}
		\vspace{-0.3cm}
	\end{center}
	\caption{Standard RoI pooling (Top) and \emph{text-alignment} layer (Bottom). Our method can avoid encoding irrelevant texts and complicated background, which is crucial for the accuracy of text recognition.}
	\label{fig:roiexamples}
	\vspace{-0.5cm}
\end{figure*}

\textbf{Network architecture} 
Our model is a fully convolutional architecture inspired by \cite{Zhou2017}, where a PVA network \cite{Hong2016} is utilized as backbone due to its significantly low computational cost. Unlike generic objects, texts often have a much larger variations in both sizes and aspect ratios. Thus it not only needs to preserve local details for small-scale text instances, but also should maintain a large receptive field for very long instances. Inspired by the success in semantic segmentation \cite{Ronneberger2015}, we exploit feature fusion by combining convolutional features of conv5, conv4, conv3 and conv2 layers gradually, with the goal of maintaining both local detailed features and high-level context information. This results in more reliable predictions on multi-scale text instances. Size of the top layer is $\frac{1}{4}$ of the input image for simplicity.

\textbf{Text detection} 
This branch is similar to that of \cite{Zhou2017}, where a multi-task prediction is implemented at each spatial location on the top convolutional maps, by adopting an Intersection over Union (IoU) loss described in \cite{Yu2016}. It contains two sub-branches on the top convolutional layer designed for joint text/non-text classification and multi-orientation bounding boxes regression.
The first sub-branch returns a classification map with an equal spatial size of the top feature maps, indicating the predicted text/non-text probabilities using a softmax function. The second sub-branch outputs five localization maps with the same spatial size, which estimate five parameters for each bounding box with arbitrary orientation at each spatial location of text regions. The five parameters represent the distances of the current point to the top, bottom, left and right sides of an associated bounding box, together with its inclined orientation. With these configurations, the detection branch is able to predict a quadrilateral of arbitrary orientation for each text instance. The feature of the detected quadrilateral region is then feed into the RNN branch for word recognition via a \emph{text-alignment} layer which is described below.

\begin{figure*}
	\begin{center}
		\includegraphics[width=13.5cm]{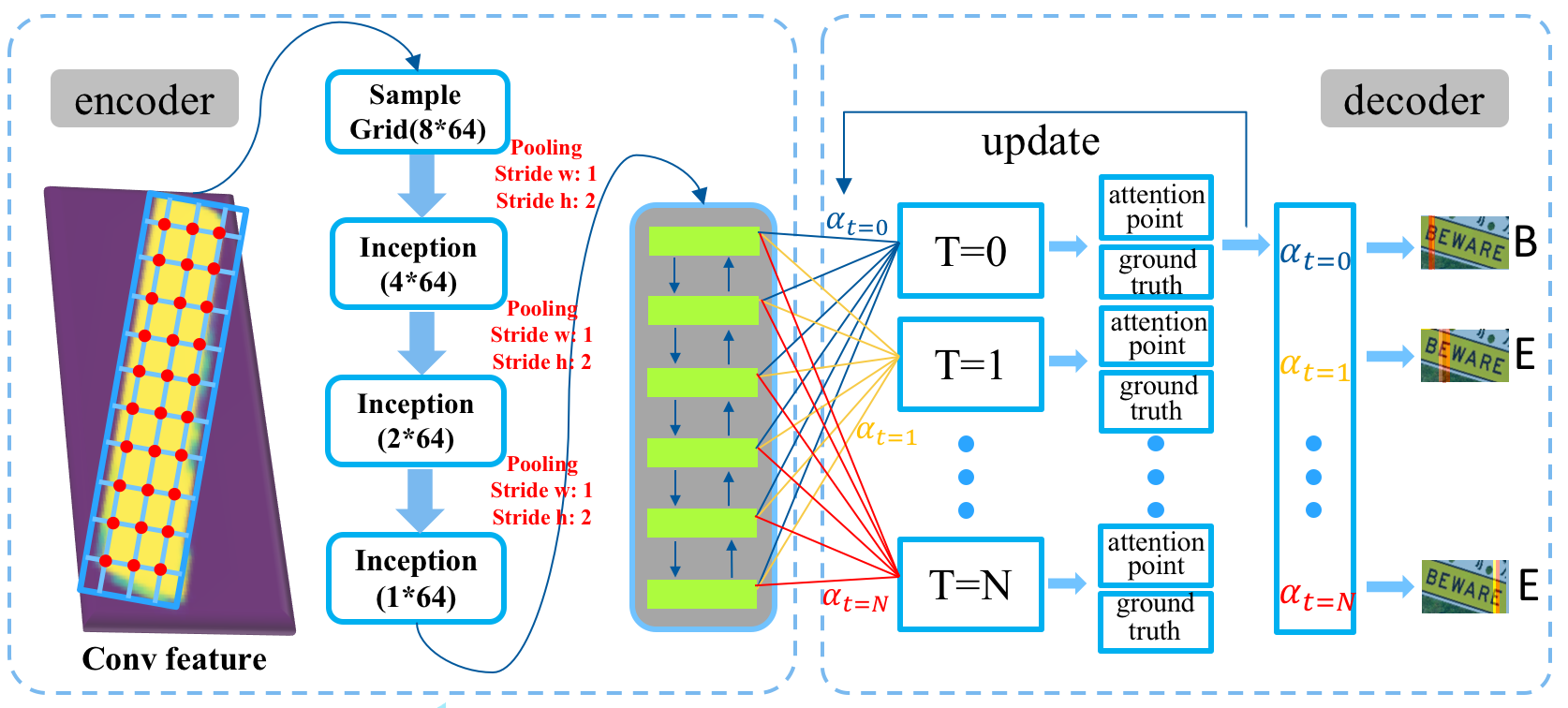}
	\end{center}
	\caption{Our proposed sub-net structure for recognition branch, which provides attention guidance during the decoding process by using character spatial information as supervision.}
	\label{fig:recogbranch}
	\vspace{-0.3cm}
\end{figure*}
\subsection{Text-Alignment Layer}
We create a new recurrent branch for word recognition, where a text-alignment layer is proposed to precisely compute fixed-size convolutional features from a quadrilateral region of arbitrary size.
The text-alignment layer is extended from RoI pooling \cite{Ross2015} which is widely used for general objects detection. The RoI pooling computes a fixed-size convolutional features (e.g., $7\times7$) from a rectangle region of arbitrary size, by performing quantization operation. It can be integrated into the convolutional layers for in-network region cropping, which is a key component for end-to-end training a detection framework. However, directly applying the RoI pooling to a text region will lead to a significant performance drop in word recognition due to the issue of misalignment.

\begin{itemize}
	\vspace{-0.15cm}
	\item[--] First, unlike object detection and classification where the RoI pooling computes global features of a RoI region for discriminating an object, word recognition requires more detailed and accurate local features and spatial information for predicting each character sequentially. As pointed out in \cite{HeK2017}, the RoI pooling performs quantizations which inevitably introduce misalignments between the original RoI region and the extracted features. Such misalignments have a significant negative effect on predicting characters, particularly on some small-scale ones such as `i', `l'.
	\vspace{-0.25cm}
	\item[--] Second, RoI pooling was designed for a rectangle region which is only capable of localizing horizontal instances. It will make larger misalignments when applied to multi-orientation text instances. Furthermore, a large amount of background information and irrelevant texts are easily encoded when a rectangle RoI region is applied to a highly inclined text instance, as shown in Fig. \ref{fig:roiexamples}. This severely reduces the performance on RNN decoding process for recognizing sequential characters.
	\vspace{-0.15cm}
\end{itemize}

Recent Mask R-CNN considers explicit per-pixel spatial correspondence by introducing RoIAlign pooling \cite{HeK2017}. This inspires current work that develops a new text-alignment layer tailored for text instance which is a quadrilateral shape with arbitrary orientation. It provides strong word-level alignment with accurate per-pixel correspondence, which is of critical importance to extract exact text information from the convolutional maps, as shown in Fig. \ref{fig:roiexamples}.

Specifically, given a quadrilateral region, we first build a sampling grid with size of $h \times w$ on the top convolutional maps. The sampled points are generated with equidistant interval within the region, and the feature vector ($\textbf{v}_p$) for a sampled point ($p$) at spatial location $(p_{x}, p_{y})$, is calculated via a bilinear sampling \cite{HeK2017} as follows,
\begin{equation}
\vspace{-0.15cm}
\label{eq:bilinear}
\textbf{v}_{p} = \sum_{i=1}^{4}{\textbf{v}_{pi} * g(p_{x}, p_{ix}) * g(p_{y}, p_{iy})}
\end{equation}
Where $\textbf{v}_{pi}$ refers to four surrounding points of point $p$, $g(m,n)$ is the bilinear interpolation function and $p_{ix}$ and $p_{iy}$ refer to the coordinates of point $p_{i}$. As presented in \cite{HeK2017}, an appealing property of the bilinear sampling is that gradients of the sampled points can be back-propagated through the networks, by using Eq. \ref{eq:bilinear_back}.
\begin{equation}
\label{eq:bilinear_back}
\frac{\partial grad}{\partial v_{pi}} = \sum{g(p_{x}, p_{ix}) * g(p_{y}, p_{iy})}
\end{equation}

Grid sampling, by generating a fixed number of sampling points (e.g., $w=64$, $h=8$ in our experiments), provides an efficient way to compute fixed-size features from a quadrilateral region with arbitrary size and orientation. The bilinear sampling allows for exacting per-pixel alignment, successfully avoiding the quantization factor.

\subsection{Word Recognition with Character Attention}
Word recognition module is built on the text-alignment layer, as shown in Fig. \ref{fig:net}. Details of this module is presented in Fig. \ref{fig:recogbranch}, where the input is fixed-size convolutional features output from the text-align pooling layer with size of $w \times h \times C$, where $C$ is the number of convolutional channels.
The convolutional features are fed into multiple inception modules and generate a sequence of feature vectors,  e.g., 64 $\times$C-dimensional features, as shown in Fig. \ref{fig:recogbranch}. In the next part, we will briefly introduce attention mechanism and three strategies to enhance attention alignment.
%

%

%\begin{figure*}[h!]
\begin{SCfigure*}[.7]
    \includegraphics[width=12.5cm]{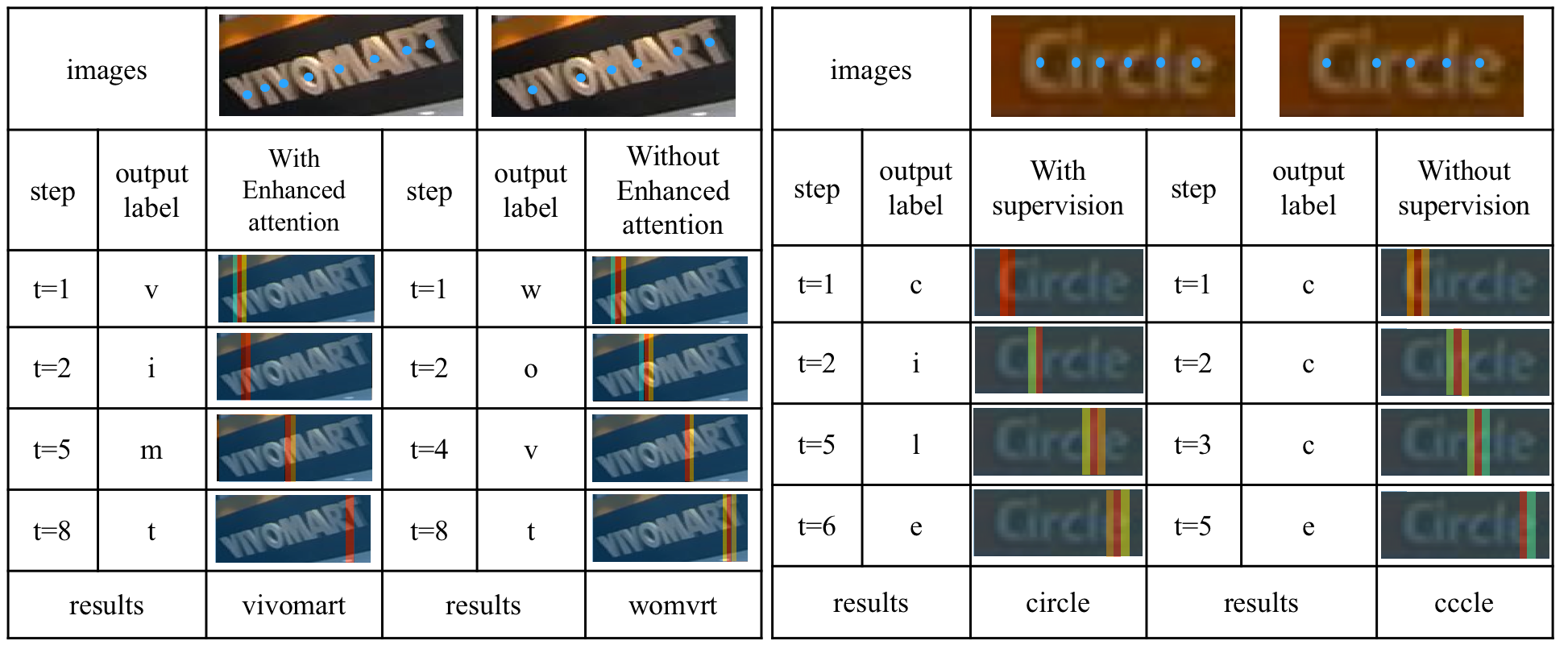}
	\caption{A comparison of the proposed method with traditional attention LSTM. The heat map indicates the focusing location at each time step.
    ~ ~ ~ ~ ~ ~ ~ ~ ~ ~ 
    ~ ~ ~ ~ ~ ~ ~ ~ ~ ~
    ~ ~ ~ ~ ~ ~ ~ ~ ~ ~
    ~ ~ ~ ~ ~ ~ ~ ~ ~ ~
    ~ ~ ~ ~ ~ ~ ~ ~ ~ ~
    ~ ~ ~ ~ ~ ~ ~ ~ ~ ~
    ~ ~ ~ ~ ~ ~ ~ ~ ~ ~
    ~ ~ ~ ~ ~ ~ ~ ~ ~ ~
    ~ ~ ~ ~ ~ ~ ~ ~ ~ ~
    ~ ~ ~ ~ ~ ~ ~ ~ ~ ~
    ~ ~ ~ ~ ~ ~ ~ ~ ~ ~
    }
	\label{fig:att_examples}
\end{SCfigure*}
%\end{figure*}

\subsubsection{Attention Mechanism}
\vspace{-0.15cm}
Recently, attention mechanism has been developed for word recognition \cite{Lee2016, Bahdanau2016}, where an implicit attention is learned automatically to enhance deep features in decoding process. In the encoding process, a bi-direction LSTM layer is utilized to encode the sequential vectors. It outputs hidden states $\{h^e_{1}, h^e_{2}, ..., h^e_{w}\}$ of the same number, which encode strong sequential context features from both past and future information.
Unlike previous work \cite{Shi2017, HeP2016} which decode a character (including a non-character label) using each hidden state, the attention mechanism introduces a new decoding process where an attention weights ($\alpha_t \in \mathbb{R}^w$) is learned automatically at each decoding iteration, and the decoder predicts a character label ($y_t$) by using this attention vector,
\begin{equation}
\vspace{-0.1cm}
\label{eq:att_lstm1}
y_{t} = Decoder(h^d_{t}, g_{t},y_{t-1})
\end{equation}
where $h^d_{t}$ is the hidden state vector of the decoder at time $t$, computed by:
\begin{equation}
\vspace{-0.15cm}
\label{eq:att_lstm2}
h^d_{t} = f(y_{t-1}, h^d_{t-1}, g_{t})
\end{equation}
$g_{t}$ is the context vector, which is calculated as a weighted sum of the input sequence: $g_{t} = \sum_{j=1}^{w}\alpha_{t, j} h^e_{j}$. The decoder is ended until it encounters an end-of-sequence ($EOS$).
The attention vector is calculated by $\alpha_{t,j} = softmax(e_{t,j})$, where $e_{t,j} = z(h^d_{t-1}, h^e_{j})$ is an alignment factor measuring matching similarity between the hidden state and encoding features $h^e_{j}$.
However, these attention vectors are learned automatically in the training process without an explicit guidance, giving rise to misalignment problem which severely reduces recognition performance, as shown in Fig.\ref{fig:att_examples}. To address this problem, we propose new attention alignment and enhancement methods that explicitly encode strong attention of each character.

\subsubsection{Attention Alignment and Enhancement}
\vspace{-0.15cm}
We introduce a new method which enhance the attention of characters in word recognition. We develop character-alignment mechanism that explicitly encodes strong character information, together with a mask supervision task which provides meaningful local details and spatial information of character for model learning. Besides, an attention position embedding is also presented. It identifies the most significant spot from the input sequence which further enhances the corresponding text features in inference. These technical improvements are integrated seamlessly into a unified framework that is end-to-end trainable. Details of each module are described as follows.

\textbf{Attention alignment} To deal with misalignment issue raised by existing implicit attention models, we propose an attention alignment which explicitly encodes spatial information of characters, by introducing an additional loss as supervision.

Specifically, assuming that $p_{t,1}, p_{t,2},..., p_{t,w}$ are central points in each column of the sampling grid. At $t$-th time step, these central points can be calculated by Eq. \ref{eq:att_pts},
\begin{equation}
\label{eq:att_pts}
\delta_{t} = \sum_{j=1}^{w}\alpha_{t, j} \times p_{t,j}
\vspace{-0.2cm}
\end{equation}
Ideally, $\delta_{t}$ should close to the center of current character, $y_{t}$.
Without supervision, it is likely to result in misalignment and therefore incorrect sequence labels.
Intuitively, we can construct a loss function to describe whether the attention points is focusing on the right location.
\begin{equation}
\vspace{-0.15cm}
\label{eq:loss_att}
\ell_{align} = \sum_{t=0}^{T} \left \| \frac{\delta_{t} - k_t}{0.5*\bar{w}_{t}}\right\|^{2}
\end{equation}
where $k_t$ is ground truth (GT) coordinates, and $\bar{w}_{t}$ is the GT width of current character, $y_{t}$. Both of them are projected onto the axis of text orientation. $T$ is the number of characters in a sequence. Notice that the distance between the prediction and GT should be normalized by character width, which we found is useful for model convergence.

\textbf{Character mask} To further enhance character attention, we introduce another additional supervision by leveraging character mask, which provides more meaningful information, including both local details and spatial location of a character. A set of binary masks are generated, with a same spatial size of the last convolutional maps. The number of the masks is equal the number of character labels. A softmax loss function is applied at each spatial location, which is referred as mask loss $\ell_{mask}$. This explicitly encoding strong detailed information of characters into the attention module. Both $\ell_{mask}$ and $\ell_{align}$ losses are optional during the training process, and can be ignored on those images where character level annotations are not provided.

\textbf{Position embedding} Position embedding was first introduced in \cite{Wojna2017}, aiming to make the model `location aware' by encoding a one-hot coordinate vector. This is equivalent to adding a varying bias terms. It is difficult to directly apply it to our task, as the size of the feature maps changes according to the size of input image. Instead, we generate a one-hot vector from the attention vector, $u_k=\mathop{\arg\min}_{j} \alpha_{t,j}$, which is a fixed-size binary vector (e.g., 64-D). Then, we directly concatenate the one-shot vector with the context vector ($g_t$), which forms a new feature representation with additional one-hot attention information. Then the decoder computed in Eq. \ref{eq:att_lstm1} can be modified as,
\begin{equation}
\label{eq:ps_embedding}
y_{t} = Decoder(h^d_{t}, g_{t}, y_{t-1}, u_{t})
\vspace{-0.2cm}
\end{equation}

Finally, by integrating all these modules into a single model, we obtains an overall loss function including four components,
\begin{equation}
\label{eq:loss_function}
L = \ell_{loc} + \ell_{word} + \lambda_{1}\ell_{align} + \lambda_{2}\ell_{mask}
\vspace{-0.2cm}
\end{equation}
where $\ell_{word}$ is a softmax loss for word recognition, $\ell_{loc}$ is the loss function for text instance detection, and $\lambda_{1}$ and $\lambda_{2}$ are corresponding loss weights (both are set to 0.1 in our experiment).

\subsection{Training Strategy}
\vspace{-0.15cm}
Training our model in an end-to-end manner is challenging due to a number of difficulties. First, largely different nature of them, e.g., significant differences in learning difficulties and convergence rates. Second, the extremely unbalanced distribution of image data. Our methods require character-level bounding boxes for generating character coordinates and masks. These detailed character annotations are not provided in the standard benchmarks, such as the ICDAR2013 \cite{ICDAR2013} and ICDAR2015 \cite{ICDAR2015}. Although Gupta~\etal \cite{Gupta2016} developed a fast and scalable engine to generate synthetic images of text, providing both word-level and character-level informations, there is still a large gap between realistic and synthesized images, making the trained model difficult to generalize well to real-world images.

We fill this gap by developing a principled training strategy which includes multiple steps. It is able to train multiple tasks collaboratively in our single model, allowing for excellent generalization capability from the synthesized images to real-world data.

\textbf{Step One}: We randomly select 600k images from the 800k synthetic images. Word recognition task is firstly trained by fixing the detection branch. We provide the ground truth (GT) bounding boxes of word instances to the text-align layer. Three losses: $\ell_{word}$, $\ell_{align} $  and $\ell_{mask}$ are computed. The training process last 120k iterations with a learning rate $2 \times 10^{-3}$.

\textbf{Step Two}: For the next 80k iterations, we open the detection branch, but still use the GT bounding boxes for the text-align layer, as the detector performs poorly at first, which will be harmful to the already trained recognition branch. The learning rate is set to $2 \times 10^{-4}$. During the next 20k iterations, sampling grid are generated from the detection branch. The model is trained end-to-end in this stage.

\textbf{Step Three}: About 3,000 real-world images from the ICDAR 2013 \cite{ICDAR2013}, ICDAR 2015 \cite{ICDAR2015} and Multi-lingual\footnote{
  \url{http://rrc.cvc.uab.es/?ch=8&com=introduction}
}
datasets are utilized in the next 60k iterations. To enhance generalization ability, data augmentation is employed. We re-scale the images by keeping aspect ratio unchanged, followed by random rotation ranging from $-20^{\circ}$ to $20^{\circ}$, and random cropping 800$\times$800 patches for training. To utilize the character-level supervision, we set the batch size to 4, where an image from synthetic dataset is included. The learning rate remained at $2 \times 10^{-4}$. The whole system is implemented by Caffe \cite{Jia2014}, with TITAN X GPUs.

\vspace{-0.25cm}

\section{Experiments}
\vspace{-0.2cm}
In this section, we first briefly introduce the datasets we use and the evaluation protocols, followed by thorough comparison of the proposed method with the state-of-the-art along with comprehensive ablation experiments.

\begin{figure*}
	\begin{center}
        \includegraphics[width=.8\textwidth]{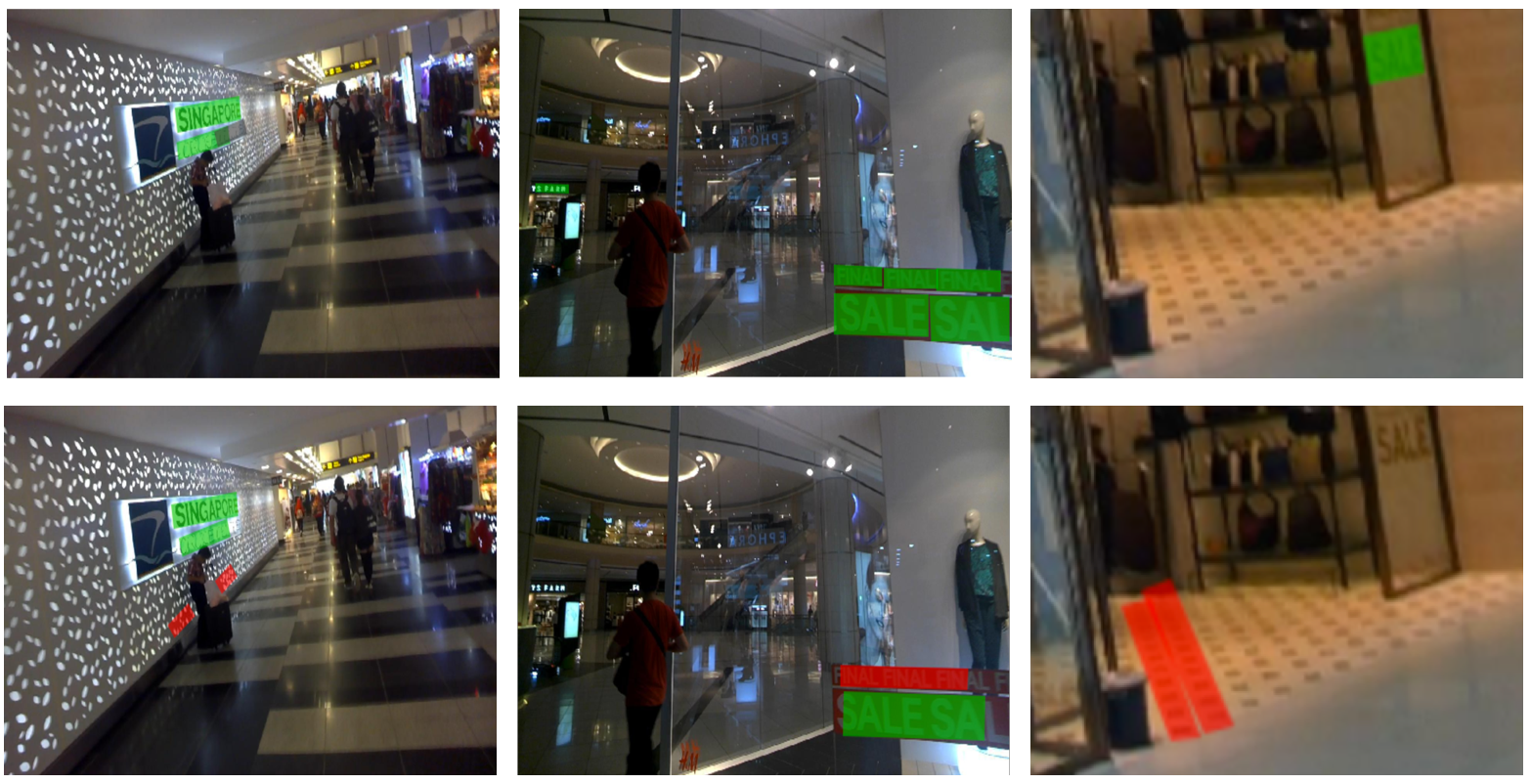}
	\end{center}
	\caption{A comparison of detection performance between joint training (Top) and separate training (Bottom). Joint training makes it more robust to find out text regions as two tasks are highly correlated, where detection can benefit from training of recognition.}
	\label{fig:example_supervise}
	\vspace{-0.3cm}
\end{figure*}

\begin{table*}[t!]
	\newcommand{\tabincell}[2]{\begin{tabular}{@{}#1@{}}#2\end{tabular}}
	\centering
	\renewcommand\arraystretch{1.3}
	\fontsize{9}{9}\selectfont
	\caption{\it Comparisons of the 
    end-to-end task with state-of-the-art on ICDAR2013 and ICDAR2015. The results are reported with three different level lexicons, namely, strong, weak and generic. }
	\label{tab:end-to-end results}
	 \setlength{\tabcolsep}{10pt}
	\begin{tabular}{c|c|c|c|c|c|c|c|c}
		\hline
		\multirow{10}{*}{\rotatebox{90}{ICDAR2013}} &
		\multirow{2}{*}{Method}&
		\multirow{2}{*}{Year} &
		\multicolumn{3}{c|}{\tabincell{c}{Word-Spotting}}&\multicolumn{3}{|c}{ \tabincell{c}{End-to-end}}\cr\cline{4-9}  & & &Strong &Weak &Generic &Strong &Weak &Generic\cr \cline{2-9}

		&Deep2Text II+ \cite{Yin2014}&2014 & $0.85$ & $0.83$ & $0.79$ & $0.82$ & $0.79$ & $0.77$\cr\cline{2-9}
		&Jaderberg~\etal \cite{Jaderberg2016b} &2015 & $0.90$ & $-$ & $0.76$ & $0.86$ & $-$ & $-$ \cr\cline{2-9}
		&FCRNall+multi-filt \cite{Gupta2016}&2016 & $-$ & $-$ & $0.85$ & $-$ & $-$ & $-$ \cr\cline{2-9}
		&TextBoxes \cite{Liao2017} &2017 & $0.94$ & $0.92$ & $0.86$ & $0.92$ & $0.90$ & $0.84$ \cr\cline{2-9}
		&YunosRobot$1.0$ &2017 & $0.87$ & $-$ & $0.87$ & $0.84$ & $-$ & $0.84$  \cr\cline{2-9}
		&Li~\etal \cite{Li2017} &2017 & $\textbf{0.94}$ & $0.92$ & $\textbf{0.88}$ & $0.91$ & $\textbf{0.90}$ & $0.85$  \cr\cline{2-9}
		&Deep text spotter \cite{Busta2017} &2017 &$0.92$ & $0.89$ & $0.81$ & $0.89$ & $0.86$ & $0.77$   \cr\cline{2-9}
		&\textbf{Proposed Method} &- &$0.93$ & $\textbf{0.92}$ & $0.87$ & $\textbf{0.91}$ & $0.89$ & $\textbf{0.86}$  \cr\cline{2-9}
		\hline
		\multirow{6}{*}{\rotatebox{90}{ICDAR2015}} &
		\multirow{2}{*}{Method}&
		\multirow{2}{*}{Year} &
		\multicolumn{3}{c|}{\tabincell{c}{Word-Spotting}}&\multicolumn{3}{|c}{ \tabincell{c}{End-to-end}}\cr\cline{4-9}  & & &Strong &Weak &Generic &Strong &Weak &Generic\cr  \cline{2-9}
		&Stradvision \cite{ICDAR2015} &2013 & $0.46$ & $-$ & $-$ & $0.44$ & $-$ & $-$ \cr \cline{2-9}
		&TextSpotter \cite{Neumann2016} &2016 & $0.37$ & $0.21$ & $0.16$ & $0.35$ & $0.20$ & $0.16$\cr \cline{2-9}
		&Deep TextSpotter \cite{Busta2017}&2017 & $0.58$ & $0.53$ & $0.51$ & $0.54$ & $0.51$ & $0.47$ \cr \cline{2-9}
		&\textbf{Proposed Method} &-  & $\textbf{0.85}$ & $\textbf{0.80}$ & $\textbf{0.65}$ &$\textbf{0.82}$ & $\textbf{0.77}$ & $\textbf{0.63}$ \cr\hline

	\end{tabular}
\vspace{-5mm}
\end{table*}

\vspace{-0.1cm}
\textbf{Datasets}
\vspace{-0.1cm}
The ICDAR2013 dataset focuses more on horizontal text instances, which contains 229 images for training and 233 images for testing with word-level annotation.

The ICDAR2015 dataset is collected by Google glasses, which has 1,000 images for training and 500 images for testing. Different from previous datasets which are well-captured horizontal English text, it contains texts with more scales, blurring, and orientation.

Multi-lingual scene text dataset\footnote{\url{http://rrc.cvc.uab.es/?ch=8&com=introduction } }
is built for developing script-robust text detection methods, which contains about 9,000 images with 9 different kinds of transcriptions. We choose about 2000 of them, identified with `Latin', to train the end-to-end task.
\begin{table*}[tp]
	\small
	\newcommand{\tabincell}[2]{\begin{tabular}{@{}#1@{}}#2\end{tabular}}
	\caption{\it Ablations for the 
    proposed method. We test our model on ICDAR2015. The
   detection part is replaced with ground truth for fair comparison.
   }
	\label{tab:contrast experiment}
    \centering
	\begin{tabular}{c|c|c|c|c|c}
		\textbf{roi pooling?}\hspace{-0.15cm} &\hspace{-0.15cm}\textbf{\tabincell{c}{roi\\alignment?}}\hspace{-0.2cm}
		&\hspace{-0.2cm} \scriptsize\textbf{\tabincell{c}{text\\alignment?}}\hspace{-0.15cm}
		&\hspace{-0.15cm}\scriptsize\textbf{supervision?}\hspace{-0.15cm}
		&\hspace{-0.15cm}\scriptsize\textbf{\tabincell{c}{ position \\ embedding?}}\hspace{-0.15cm}
		&\hspace{-0.15cm}\scriptsize\textbf{\tabincell{c}{ Accuracy (\%)} } \cr

		$\checkmark$ &$\times$  &$\times$ & $\times$ &$\times$ &60.7 \cr
		$\times$ &$\checkmark$  &$\times$ & $\times$ &$\times$ &61.9 \cr
		$\times$ &$\times$ &$\checkmark$ & $\times$ &$\times$ &67.6 \cr
		$\times$ &$\times$ &$\checkmark$ & $\checkmark$ &$\times$ &68.8 \cr
		$\times$ &$\times$ &$\checkmark$ & $\times$ &$\checkmark$ &68.2 \cr
		$\times$ &$\times$ &$\checkmark$ & $\checkmark$ &$\checkmark$ &\textbf{69.5} \cr

	\end{tabular}
	\vspace{-7mm}
\end{table*}

\begin{table*}[tp]
	\newcommand{\tabincell}[2]{\begin{tabular}{@{}#1@{}}#2\end{tabular}}
	\centering
	\renewcommand\arraystretch{1.3}
	\fontsize{8}{8}\selectfont
	\caption{\it Comparison of detection results with the state-of-the-art methods on ICDAR2013 and ICDAR2015. The results are reported
  Recall (R), Precision (P) and F-measure (F). For fair comparison, the detection performance is achieved without referring to recognition results.}
	\label{tab:detection results}
	\begin{tabular}{l|c|c|c|c|c|c|c|l|c|c|c|c}
		\hline
		\multicolumn{8}{c}{ICDAR2013 dataset} &
		\multicolumn{5}{c}{ICDAR2015 dataset} \\
		\hline
		\hline
		\multirow{2}{*}{Method} & \multirow{2}{*}{Year} &
		\multicolumn{3}{c|}{ICDAR standard} &\multicolumn{3}{c|}{DetEval} & \multirow{2}{*}{Method} & \multirow{2}{*}{Year} & \multirow{2}{*}{R} & \multirow{2}{*}{P} & \multirow{2}{*}{F} \\
		\cline{3-8}
		& &R &P &F &R &P &F & & & & &
		\\
		\hline
		TextFlow \cite{Tian2015} &2015 &0.76 &0.85 &0.80 &- &- &- &StradVision2 &2015 &0.37 &0.77 &0.50 \cr\hline
		Text-CNN \cite{HeT2016} &2016 &0.73 &0.93 &0.82 &0.76 &0.93 &0.84 &MCLAB\_FCN \cite{Zhang2016} &2016 &0.43 &0.71 &0.54 \cr\hline
		FCRN \cite{Gupta2016} &2016 &0.76 &\textbf{0.94} &0.84 &0.76 &0.92 &0.83 &EAST \cite{Zhou2017} &2016 &0.78 &0.83 &0.81 \cr\hline
		CTPN \cite{Tian2016} &2016 &0.73 &0.93 &0.82 &0.83 &\textbf{0.93} &0.88 &CTPN \cite{Tian2016} &2016 &0.52 & 0.74 &0.61 \cr\hline
		He~\etal \cite{HeP2017} &2017 &0.86 &0.88 &0.87 &0.86 &0.89 &0.88 & He~\etal \cite{HeP2017} &2017 &0.73 &0.80 &0.77 \cr\hline
		He~\etal \cite{HeW2017} &2017 &0.81 &0.92 &0.86 &- &- &- & He~\etal \cite{HeW2017} &2017 & 0.82 &0.80 &0.81\cr\hline\hline
		Proposed wo recog &- &0.87 &0.88 &0.88 &0.87 &0.88 &0.88 & Proposed wo recog &- &0.83 &0.84 &0.83\cr\hline
		\textbf{Proposed} &- &\textbf{0.88} &0.91 & \textbf{0.90} &\textbf{0.89} &0.91 &\textbf{0.90} &\textbf{Proposed}  &- &\textbf{0.86} &\textbf{0.87} &\textbf{0.87}\cr\hline

	\end{tabular}
\vspace{-0.5cm}
\end{table*}

\vspace{-0.15cm}
\subsection{Evaluation Protocols}
\textbf{Detection} There are two standard protocols for evaluating detection results: DetEval and
ICDAR2013 standard \cite{ICDAR2013}. The main difference between the two protocols is that the latter one stress more on individual words while the former can achieve high score even when many words are connected into a line.

\textbf{End-to-end for detection and recognition}
The criterion has been used in competition: the evaluation of the results will be based on a single IoU criterion, with a threshold of 50\%, and correct transcription.
Besides, three dictionaries are also provided for testing reference, i.e., `strong', `weak' and `generic'. `Strong' lexicon has 100 entries for every image, and most words appeared in that image are included. `Weak' lexicon contains all the words that appeared in the testing dataset. 'Generic' lexicon has 90K words. One thing should be noticed that the length of all the words in dictionaries are greater than 3 with symbols and numbers excluded.
There are two protocols for evaluation: end-to-end and word-spotting. End-to-end needs to recognize all the words precisely, no matter whether the dictionary contains these strings. On the other hand, word-spotting only examine whether the words in the dictionary appear in images, making it less strict than end-to-end for ignoring symbols, numbers and words whose length is less than 3.

\vspace{-0.25cm}
\subsection{Text-alignment vs.\ RoI Pooling}
\vspace{-0.15cm}
We first compare the proposed \emph{text-alignment} with standard RoI pooling. To make fair comparison, the detection part is fixed with ground truth and recognition performance is evaluated on ICDAR2015, which contains text instances of multi-orientation. Due to encoding background information and irrelevant text instances, RoI pooling results in mis-alignment and inaccurate representation of feature sequences. As shown in Tab. \ref{tab:contrast experiment}, the accuracy of recognition with proposed method surpasses standard RoI pooling by a large margin, boosting from 60.7\% to 67.6\%. All results are evaluated without referring to any lexicon in single scale.

\vspace{-0.3cm}
\subsection{Character Attention}
\vspace{-0.15cm}
Different from traditional attention-based recognition models, where attention weights are automatically learned, we propose a method to regulate the learning process to prevent mis-alignment in the decoding stage. To demonstrate the effectiveness of our proposed method, we conduct two experiments with the detection part fixed. The first one is on VGG synthetic data \cite{Gupta2016}, where we select 600K for training and 200K for testing. The accuracies of character-level and word-level are evaluated. The method with supervision has accuracy of 0.95 and 0.88 on two protocols, comparing to 0.93 and 0.85 on traditional attention-based method. The other experiment is tested on ICDAR2015 dataset. As is shown in Fig. \ref{fig:att_examples}, the proposed method give more accurate character localization than attentional LSTM, leading to about 2\% boosting in accuracy.

\vspace{-0.15cm}
\subsection{Joint Training vs.\ Separate Models}
\vspace{-0.15cm}
We believe that text detection and recognition are not two standalone problems, but highly correlated where each task can benefit from the training of the other. Joint training of two tasks in a unified framework avoids error accumulations among cascade models. As shown in Tab. \ref{tab:detection results}, the task of recognition greatly enhances the performance of detection in terms of recall and precision, leading to a 3\% improvement on F-Measure (noting: the detection performances are achieved without referring to recognition results). As can be seen from Fig. \ref{fig:example_supervise}, joint training makes it more robust to text-like background and complicated text instances. We also provide a comparison with other detection approaches, indicating that our method achieved new state-of-the-art performance on ICDAR2013 and ICDAR2015 datasets.

\vspace{-0.15cm}
\subsection{Proposed Method vs. State-of-the-art Methods}
\vspace{-0.15cm}
\begin{figure*}
	\begin{center}
        \includegraphics[width=0.8\textwidth]{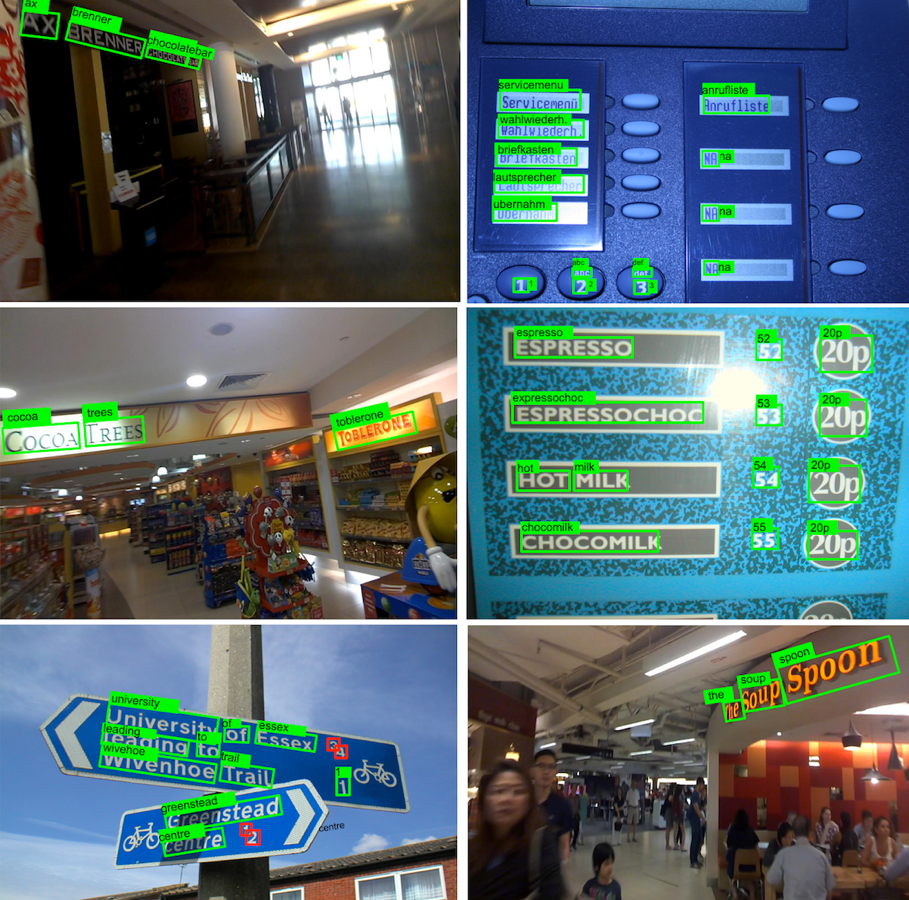}
	\end{center}
	\caption{Examples of textspotting results of the proposed method on ICDAR2013 and ICDAR2015. }
	\label{fig:results}
	\vspace{-0.5cm}
\end{figure*}

End-to-end results on some extremely challenging images are presented in Fig. \ref{fig:results}. As can be seen in Fig. \ref{fig:results}, our method can correctly detect and recognize both small text instances and those with large inclined angles.

\textbf{ ICDAR2015}  The effectiveness to multi-orientation texts is testified on ICDAR2015 dataset. Our method achieved an F-measure of 0.82, 0.77 and 0.63 respectively in terms of referencing `Strong', `Weak' and `Generic' lexicon under the end-to-end protocol, which surpasses the state-of-the-art performance of 0.54, 0.77 and 0.63 by a large margin.

\textbf{ICDAR2013} The dataset is well-captured for horizontal text instances. The result is shown in Tab. \ref{tab:end-to-end results}, which is comparable to the state-of-the-art result \cite{Li2017}.

\vspace{-0.2cm}

	\textbf{Conclusion}
\vspace{-0.2cm}
In this paper we have presented a novel framework that combines detection and recognition in a unified network with sharable features. The model can directly output detection and recognition results of multi-orientation text instances.

We have proposed a novel text-alignment layer that can extract precise sequence information without encoding irrelevant background or texts. We also improve the accuracy of traditional LSTM by enhancing the attention of characters during the decoding process.
Our proposed method achieves state-of-the-art performance on two open benchmarks: ICDAR2013 and ICDAR2015 and outperforms previous best methods by a large margin.

\textbf{Acknowledgments}
C. Shen's participation was in part supported by an ARC Future Fellowship.

	{\small
		\bibliographystyle{ieeetr}
		\bibliography{final}

\begin{thebibliography}{10}

\bibitem{ICDAR2015}
D.~Karatzas, L.~Gomez-Bigorda, A.~Nicolaou, S.~Ghosh, A.~Bagdanov, M.~Iwamura,
  J.~Matas, L.~Neumann, V.~R. Chandrasekhar, S.~Lu, F.~Shafait, S.~Uchida, and
  E.~Valveny, ``{ICDAR} 2015 competition on robust reading,'' in {\em Proc.
  Int. Conf. Document Analysis \& Recognition}, 2015.

\bibitem{Busta2017}
M.~Busta, L.~Neumann, and J.~Matas, ``Deep textspotter: An end-to-end trainable
  scene text localization and recognition framework,'' in {\em Proc. IEEE Int.
  Conf. Comp. Vis.}, 2017.

\bibitem{Jaderberg2015}
M.~Jaderberg, K.~Simonyan, A.~Zisserman, and K.~Kavukcuoglu, ``Spatial
  transformer networks,'' in {\em Proc. Advances in Neural Inf. Process.
  Syst.}, 2015.

\bibitem{HeP2017}
P.~He, W.~Huang, T.~He, Q.~Zhu, Y.~Qiao, and X.~Li, ``Single shot text detector
  with regional attention,'' in {\em Proc. IEEE Int. Conf. Comp. Vis.}, 2017.

\bibitem{Li2017}
H.~Li, P.~Wang, and C.~Shen, ``Towards end-to-end text spotting with
  convolutional recurrent neural networks,'' in {\em Proc. IEEE Int. Conf.
  Comp. Vis.}, 2017.

\bibitem{HeW2017}
W.~He, X.~Zhang, F.~Yin, and C.~Liu, ``Deep direct regression for
  multi-oriented scene text detection,'' {\em arXiv preprint arXiv:1703.08289},
  2017.

\bibitem{Tian2016}
Z.~Tian, W.~Huang, T.~He, P.~He, and Y.~Qiao, ``Detecting text in natural image
  with connectionist text proposal network,'' in {\em Proc. Eur. Conf. Comp.
  Vis.}, 2016.

\bibitem{Zhou2017}
X.~Zhou, C.~Yao, H.~Wen, Y.~Wang, S.~Zhou, W.~He, and J.~Liang, ``{EAST}: An
  efficient and accurate scene text detector,'' in {\em Proc. IEEE Conf. Comp.
  Vis. Patt. Recogn.}, 2017.

\bibitem{Shi2017}
B.~Shi, X.~Bai, and C.~Yao, ``An end-to-end trainable neural network for
  image-based sequence recognition and its application to scene text
  recognition,'' {\em {IEEE} Trans. Pattern Anal. Mach. Intell.}, vol.~39,
  pp.~2298--2304, 2017.

\bibitem{Tian2017}
S.~Tian, S.~Lu, and C.~Li, ``Wetext: Scene text detection under weak
  supervision,'' in {\em Proc. IEEE Int. Conf. Comp. Vis.}, 2017.

\bibitem{Liu2018}
X.~Liu, D.~Liang, S.~Yan, D.~Chen, Y.~Qiao, and J.~Yan, ``Fots: Fast oriented
  text spotting with a unified network,'' {\em arXiv preprint
  arXiv:1801.01671}, 2018.

\bibitem{Ren2015}
S.~Ren, K.~He, R.~Girshick, and J.~Sun, ``Faster {R-CNN}: Towards real-time
  object detection with region,'' in {\em Proc. Advances in Neural Inf.
  Process. Syst.}, 2015.

\bibitem{LiuW2016}
W.~Liu, D.~Anguelov, D.~Erhan, C.~Szegedy, S.~Reed, C.~Fu, and A.~C. Berg,
  ``{SSD}: Single shot multibox detector,'' in {\em Proc. Eur. Conf. Comp.
  Vis.}, 2016.

\bibitem{Long2015}
J.~Long, E.~Shelhamer, and T.~Darrell, ``Fully convolutional networks for
  semantic segmentation,'' in {\em Proc. IEEE Conf. Comp. Vis. Patt. Recogn.},
  2015.

\bibitem{Zhang2016}
Z.~Zhang, C.~Zhang, W.~Shen, C.~Yao, W.~Liu, and X.~Bai, ``Multi-oriented text
  detection with fully convolutional networks,'' in {\em Proc. IEEE Conf. Comp.
  Vis. Patt. Recogn.}, 2016.

\bibitem{HeP2016}
P.~He, W.~Huang, Y.~Qiao, C.~C. Loy, and X.~Tang, ``Reading scene text in deep
  convolutional sequences,'' in {\em Proc. {AAAI} Conf. Artificial Intell.},
  2016.

\bibitem{Lee2016}
C.~Lee and S.~Osindero, ``Recursive recurrent nets with attention modeling for
  {OCR} in the wild,'' in {\em Proc. IEEE Conf. Comp. Vis. Patt. Recogn.},
  2016.

\bibitem{Bahdanau2016}
D.~Bahdanau, K.~Cho, and Y.~Bengio, ``Neural machine translation by jointly
  learning to align and translate,'' {\em arXiv preprint arXiv:1409.0473},
  2016.

\bibitem{HeK2017}
K.~He, G.~Gkioxari, P.~Dollar, and R.~Grishick, ``Mask {R-CNN},'' in {\em Proc.
  IEEE Int. Conf. Comp. Vis.}, 2017.

\bibitem{Gupta2016}
A.~Gupta, A.~Vedaldi, and A.~Zisserman, ``Synthetic data for text localisation
  in natural images,'' in {\em Proc. IEEE Conf. Comp. Vis. Patt. Recogn.},
  2016.

\bibitem{Huang2013}
W.~Huang, Z.~Lin, J.~Yang, and J.~Wang, ``Text localization in natural images
  using stroke feature transform and text covariance descriptors,'' in {\em
  Proc. IEEE Int. Conf. Comp. Vis.}, 2013.

\bibitem{Huang2014}
W.~Huang, Y.~Qiao, and X.~Tang, ``Robust scene text detection with
  convolutional neural networks induced {MSER} trees,'' in {\em Proc. Eur.
  Conf. Comp. Vis.}, 2014.

\bibitem{HeT2016}
T.~He, W.~Huang, Y.~Qiao, and J.~Yao, ``Text-attentional convolutional neural
  networks for scene text detection,'' {\em {IEEE} Trans. Image Process.},
  vol.~25, pp.~2529--2541, 2016.

\bibitem{Yao2012}
C.~Yao, X.~Bai, W.~Liu, Y.~Ma, and Z.~Tu, ``Detecting texts of arbitrary
  orientations in natural images,'' in {\em Proc. IEEE Conf. Comp. Vis. Patt.
  Recogn.}, 2012.

\bibitem{Liao2017}
M.~Liao, B.~Shi, X.~Bai, X.~Wang, and W.~Liu, ``Textboxes: A fast text detector
  with a single deep neural network,'' in {\em Proc. {AAAI} Conf. Artificial
  Intell.}, 2017.

\bibitem{HeT2016b}
T.~He, W.~Huang, Y.~Qiao, and J.~Yao, ``Accurate text localization in natural
  image with cascaded convolutional text network,'' {\em arXiv preprint
  arXiv:1603.09423}, 2016.

\bibitem{Zhang2015}
Z.~Zhang, W.~Shen, C.~Yao, and X.~Bai, ``Symmetry-based text line detection in
  natural scenes,'' in {\em Proc. IEEE Conf. Comp. Vis. Patt. Recogn.}, 2015.

\bibitem{Yao2016}
C.~Yao, X.~Bai, N.~Sang, X.~Zhou, S.~Zhou, and Z.~Cao, ``Scene text detection
  via holistic, multi-channel prediction,'' {\em arXiv preprint
  arXiv:1606.09002}, 2016.

\bibitem{Shi2016}
B.~Shi, X.~Wang, P.~Lyu, C.~Yao, and X.~Bai., ``Robust scene text recognition
  with automatic rectification,'' in {\em Proc. IEEE Conf. Comp. Vis. Patt.
  Recogn.}, 2016.

\bibitem{Alex2014}
A.~Graves and N.~Jaitly, ``Towards end-to-end speech recognition with recurrent
  neural networks,'' in {\em Proc. Int. Conf. Mach. Learn.}, 2014.

\bibitem{Alex2009}
A.~Graves, M.~Liwicki, S.~Fernandez, R.~Bertolami, H.~Bunke, and
  J.~Schmidhuber, ``A novel connectionist system for unconstrained handwriting
  recognition,'' {\em {IEEE} Trans. Pattern Anal. Mach. Intell.}, vol.~31,
  pp.~855--868, 2009.

\bibitem{Hong2016}
S.~Hong, B.~Roh, K.~Kim, Y.~Cheon, and M.~Park, ``{PVANet}: Lightweight deep
  neural networks for real-time object detection,'' {\em arXiv preprint
  arXiv:1611.08588}, 2016.

\bibitem{Ronneberger2015}
O.~Ronneberger, P.~Fischer, and T.~Brox, ``{U-Net}: Convolutional networks for
  biomedical image segmentation,'' in {\em Proc. Int. Conf. Medical Image
  Computing \& Computer-Assisted Intervention}, 2015.

\bibitem{Yu2016}
J.~Yu, Y.~Jiang, Z.~Wang, Z.~Cao, and T.~Huang, ``{UnitBox}: An advanced object
  detection network,'' in {\em ACM Conf. Multimedia}, 2016.

\bibitem{Ross2015}
R.~Grishick, ``Fast {R-CNN},'' in {\em Proc. IEEE Int. Conf. Comp. Vis.}, 2015.

\bibitem{Wojna2017}
Z.~Wojna, A.~Gorban, D.~Lee, K.~Murphy, Q.~Yu, Y.~Li, and J.~Ibarz,
  ``Attention-based extraction of structured information from street view
  imagery,'' {\em arXiv preprint arXiv:1704.03549}, 2017.

\bibitem{ICDAR2013}
D.~Karatzas, F.~Shafait, S.~Uchida, M.~Iwamura, L.~Gomez, S.~Robles, J.~Mas,
  D.~Fernandez, J.~Almazan, and L.~de~las Heras, ``{ICDAR} 2013 robust reading
  competition,'' in {\em Proc. Int. Conf. Document Analysis and Recognition},
  2013.

\bibitem{Jia2014}
Y.~Jia, E.~Shelhamer, J.~Donahue, S.~Karayev, J.~Long, R.~Girshick,
  S.~Guadarrama, and T.~Darrell, ``Caffe: Convolutional architecture for fast
  feature embedding,'' in {\em ACM Conf. Multimedia}, 2014.

\bibitem{Yin2014}
X.~Yin, X.~Yin, K.~Huang, and H.~Hao, ``Robust text detection in natural scene
  images,'' {\em {IEEE} Trans. Pattern Anal. Mach. Intell.}, vol.~36,
  pp.~970--983, 2014.

\bibitem{Jaderberg2016b}
M.~Jaderberg, K.~Simonyan, A.~Vedaldi, and A.~Zisserman, ``Reading text in the
  wild with convolutional neural networks,'' {\em Int. J. Comput. Vision},
  vol.~116, pp.~1--20, 2016.

\bibitem{Neumann2016}
L.~Neumann and J.~Matas, ``Real-time lexicon-free scene text localization and
  recognition,'' {\em {IEEE} Trans. Pattern Anal. Mach. Intell.}, vol.~38,
  pp.~1872--1885, 2016.

\bibitem{Tian2015}
S.~Tian, Y.~Pan, C.~Huang, S.~Lu, K.~Yu, and C.~L. Tan, ``Text flow: A unified
  text detection system in natural scene images,'' in {\em Proc. IEEE Int.
  Conf. Comp. Vis.}, 2015.

\end{thebibliography}
	}

\end{document}